\documentclass[sigconf]{acmart}
\usepackage[shortlabels]{enumitem}
\setlist[itemize]{leftmargin=*}
\setlist[enumerate]{leftmargin=*}
\usepackage{multirow}
\usepackage{times}

\usepackage[linesnumbered,ruled,vlined]{algorithm2e}

\newtheorem{problem}{Problem}

\AtBeginDocument{%
  \providecommand\BibTeX{{%
    \normalfont B\kern-0.5em{\scshape i\kern-0.25em b}\kern-0.8em\TeX}}}

\copyrightyear{2020}
\acmYear{2020} 
\setcopyright{iw3c2w3}
\acmConference[WWW '20]{Proceedings of The Web Conference 2020}{April 20--24, 2020}{Taipei, Taiwan} 
\acmBooktitle{Proceedings of The Web Conference 2020 (WWW '20), April 20--24, 2020, Taipei, Taiwan}
\acmPrice{}
\acmDOI{10.1145/3366423.3380144}
\acmISBN{978-1-4503-7023-3/20/04}

\begin{document}

\title{Snippext: Semi-supervised Opinion Mining with Augmented~Data}

\author{Zhengjie Miao}
\authornote{This work was done during an internship at Megagon Labs}
\email{zjmiao@cs.duke.edu}
\affiliation{%
  \institution{Duke University}
}
\author{Yuliang Li}
\email{yuliang@megagon.ai}
\affiliation{%
  \institution{Megagon Labs}
}
\author{Xiaolan Wang}
\email{xiaolan@megagon.ai}
\affiliation{%
  \institution{Megagon Labs}
}
\author{Wang-Chiew Tan}
\email{wangchiew@megagon.ai}
\affiliation{%
  \institution{Megagon Labs}
}


\renewcommand{\shortauthors}{Z. Miao, Y. Li, X. Wang, and W. Tan}

\newcommand{\yuliang}[1]{{\it\small\textcolor{blue}{[[[ {#1}\ --yuliang ]]]}}}
\newcommand{\zhengjie}[1]{{\it\small\textcolor{brown}{[[[ {#1}\ --zhengjie ]]]}}}
\newcommand{\xiaolan}[1]{{\it\small\textcolor{cyan}{[[[ {#1}\ --xiaolan ]]]}}}
\newcommand{\wctan}[1]{{\it\small\textcolor{red}{[[[ {#1}\ --wangchiew ]]]}}}

\renewcommand{\yuliang}[1]{}
\renewcommand{\zhengjie}[1]{}
\renewcommand{\xiaolan}[1]{}
\renewcommand{\wctan}[1]{}

\newcommand{\attr}{\ensuremath{\mathsf{attr}}}
\newcommand{\asp}{\ensuremath{\mathsf{asp}}}
\newcommand{\opi}{\ensuremath{\mathsf{opi}}}
\newcommand{\opinedb}{\ensuremath{\mathsf{OpineDB}}}
\newcommand{\da}{\ensuremath{\mathsf{DA}}}
\newcommand{\mixup}{\ensuremath{\mathsf{MixUp}}}
\newcommand{\mixupnl}{\ensuremath{\mathsf{MixUp^{NL}}}}
\newcommand{\mixda}{\ensuremath{\mathsf{MixDA}}}
\newcommand{\mixmatch}{\ensuremath{\mathsf{MixMatch}}}
\newcommand{\mixmatchnl}{\ensuremath{\mathsf{MixMatch^{NL}}}}
\newcommand{\snippext}{\ensuremath{\mathsf{Snippext}}}

\begin{abstract}
Online services are interested in solutions to opinion mining, which is the problem of extracting aspects, opinions, and sentiments from text. 
One method to mine opinions is to leverage the recent success of pre-trained language models which can be fine-tuned to obtain high-quality extractions from reviews. However, fine-tuning language models still requires a non-trivial amount of training data.

In this paper, we study the problem of how to significantly 
reduce the amount of labeled training data required in fine-tuning language models for opinion mining. 
We describe \snippext, an opinion mining system developed over a language model 
that is fine-tuned through semi-supervised learning with augmented data. 
A novelty of \snippext\ is its clever use of a two-prong approach to 
achieve state-of-the-art (SOTA) performance with little labeled training data 
through: (1) data augmentation to automatically generate more labeled training data from existing ones, and (2) a semi-supervised learning technique to leverage 
the massive amount of unlabeled data in addition to the (limited amount of) labeled data. 
We show with extensive experiments that \snippext\ performs comparably and can even exceed previous SOTA results
on several opinion mining tasks with only half the training data required. Furthermore, it 
achieves new SOTA results when all training data are leveraged.
By comparison to a baseline pipeline, 
we found that \snippext\ extracts significantly more fine-grained opinions which enable new opportunities of downstream applications.
\end{abstract}
\maketitle

\ccsdesc[500]{Information systems~Sentiment analysis}
\keywords{Sentiment Analysis, Semi-supervised Learning, Data augmentation, MixUp, Fine-tuning Language Models}

\vspace{-2mm}
\section{Introduction}
\label{sec:intro}


Online services such as Amazon, Yelp, or Booking.com are constantly extracting
aspects, opinions, and sentiments from reviews and other online sources of user-generated information. Such extractions are useful for obtaining
insights about services, consumers, or products and answering 
consumer questions. Aggregating the extractions can also provide summaries of 
actual user experiences directly to consumers so that they do not have
to peruse all reviews or other sources of information.
One method to easily mine opinions with a 
good degree of accuracy
is to leverage the success of pre-trained language models such 
as BERT \cite{bert} or XLNet \cite{xlnet} which can be fine-tuned to 
obtain high-quality extractions from text.
However, fine-tuning language
models still requires a significant amount of high-quality labeled training data.
Such labeled training data are usually expensive and time-consuming to
obtain as they often involve a great amount of human effort.
Hence, there has been significant research interest in
obtaining quality labeled data in a less expensive or more
efficient way \cite{settles2008active,settles2009active}.

In this paper, we study the problem of how to reduce the amount of
labeled training data required in fine-tuning language models for opinion
mining. We describe \snippext, an opinion mining system developed
based on a language model that is fine-tuned through semi-supervised
learning with augmented data. \snippext\ is motivated by the need to
accurately mine opinions, with small amounts of labeled training data,
from reviews of different domains, such as hotels, restaurants,
companies, etc.

\begin{example}
\label{exm:extractor}
\snippext\ mines three main types of information from reviews: 
{\em aspects}, {\em opinions}, and {\em sentiments}, which the following example illustrates. 

\begin{figure}[ht!]
\vspace{-3mm}
    \centering
    \includegraphics[width=0.48\textwidth]{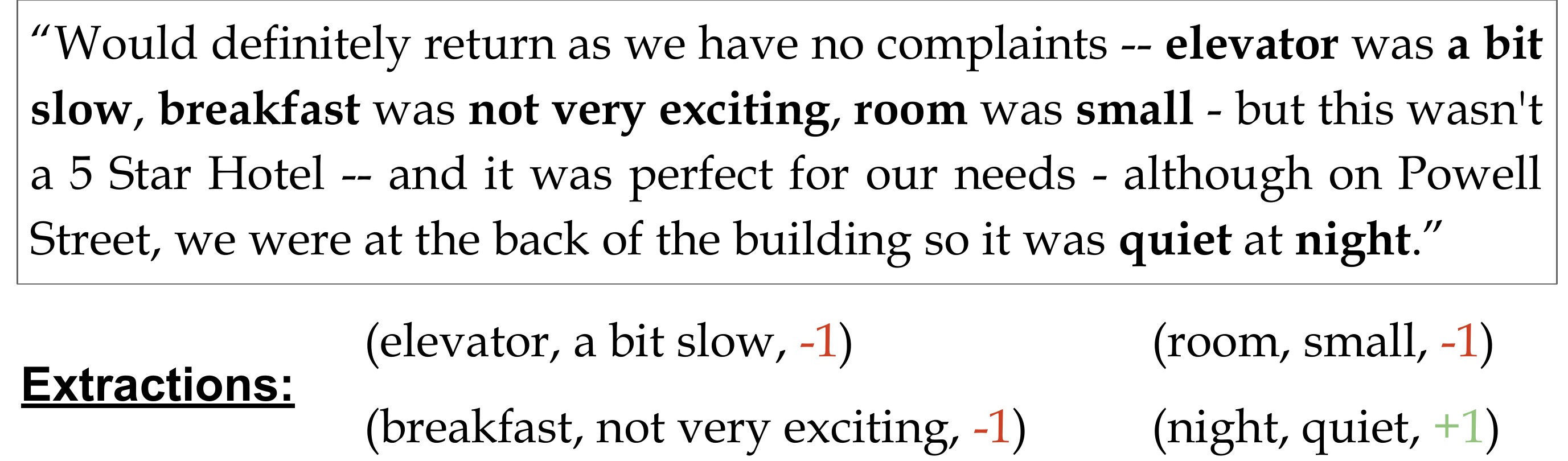}
\vspace{-6mm}
    \caption{\small{Extractions from a hotel review.}}
\vspace{-3mm}
    \label{fig:extractor}
\end{figure}

Figure \ref{fig:extractor} shows an example where triples of
the form $(\asp, \opi, s)$ are derived from a hotel review. 
For example, the triple 
(elevator, a bit slow, -1) consists of two spans of tokens that are extracted from the review, where 
``a bit slow'' is an {\em opinion term} about the {\em aspect term} ``elevator''. 
The polarity score -1 is derived based on the sentence that contains the aspect and opinion terms and it indicates a negative sentiment in this example. (1 indicates positive, -1 is negative, and 0 is neutral.) 
\end{example}

\begin{figure*}[!ht]
    \centering
    \includegraphics[width=0.65\textwidth]{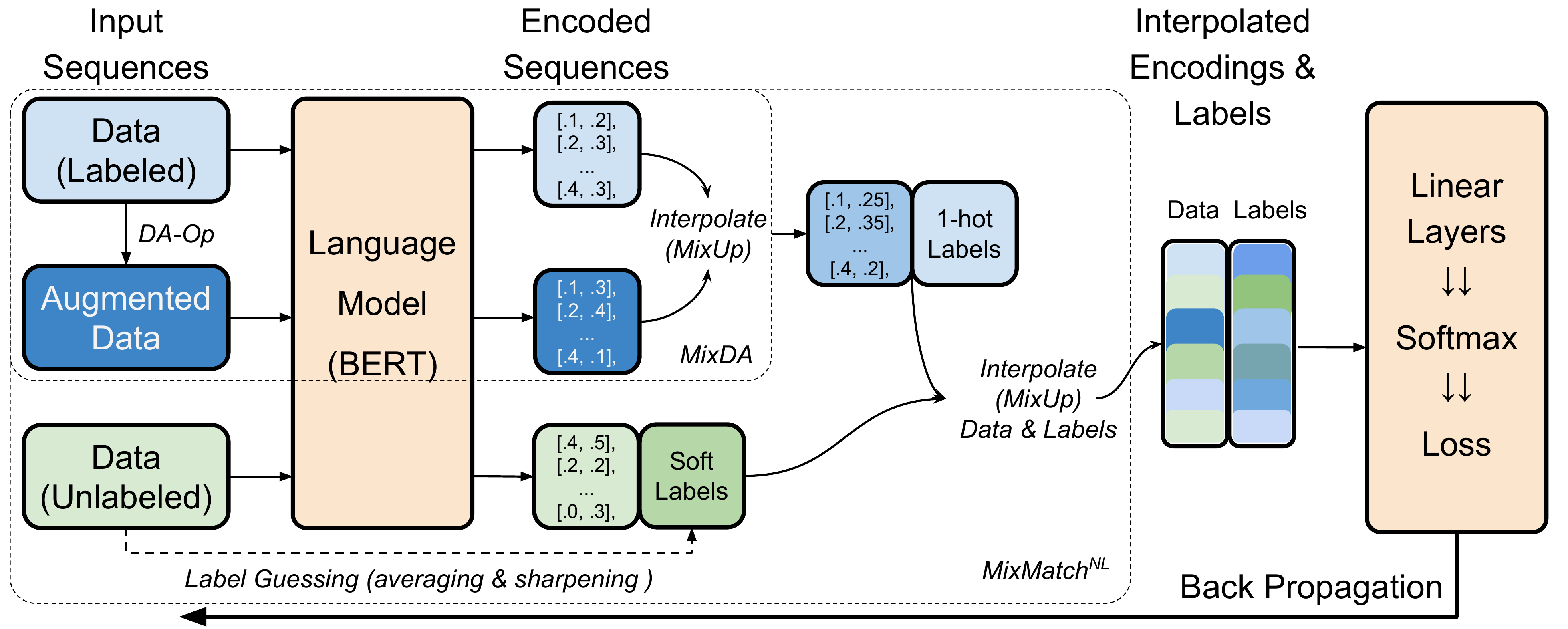}
\vspace{-4mm}
    \caption{\small Overall Architecture of \snippext. \mixda\ augments and interpolate
    the labeled training sequences. \mixmatchnl\ further combines labeled and unlabeled
    data into supervisory signal for fine-tuning the pre-trained Language Model (LM). }
\vspace{-4mm}
    \label{fig:snippext}
\end{figure*}


\vspace{-2mm}
As mentioned earlier, 
one can simply fine-tune a pre-trained language model such as BERT \cite{bert} using labeled training data to obtain
the triples as shown in Figure~\ref{fig:extractor}.
Recent results \cite{posttrain,auxiliarysentence,opinedb} showed
that BERT with fine-tuning achieved state-of-the-art (SOTA) performance in many extraction tasks,
outperforming previous customized neural network approaches. 
However, the fine-tuning approach still requires a significant
amount of high-quality labeled training data.
For example, the SOTA for aspect term extraction for restaurants is trained on
3,841 sentences labeled by linguistic experts through a non-trivial
process \cite{pontiki2014semeval}. 
In many cases, labeled training data are obtained by crowdsourcing \cite{li2016crowdsourced}.
Even if the monetary cost of crowdsourcing may not be an issue, 
preparing crowdsourcing task, launching, 
and post-processing the results are usually very
time-consuming. The process often needs to be repeated a few times to
make necessary adjustments. Also, in many cases, measures have to
be taken to remove malicious crowdworkers and to ensure the quality of 
crowdworkers.
Furthermore, the labels for a
sentence have to be collected several times to reduce possible errors
and the results have to be cleaned before they are consumable for
downstream tasks.
Even worse, this expensive labeling process has to be repeated to
train the model on each different domain (e.g., company reviews).

Motivated by the aforementioned issues, we investigate the problem of reducing the amount of labeled training
data required for fine-tuning language models such as BERT.
Specifically, we investigate solutions to the following problem.

\vspace{-1.5mm}
\begin{problem}
Given the problem of extracting aspect and opinion pairs from reviews, 
and deriving the corresponding sentiment of each aspect-opinion pair,
can we fine-tune a language model with half (or less) of training examples 
and still perform comparably with the SOTA? 
\end{problem}
\vspace{-1.5mm}


\noindent
{\bf Contributions~} 
We present \snippext, our solution to the above problem. The architecture
of \snippext\ is depicted in~Figure~\ref{fig:snippext}. 
Specifically, we make the following contributions:

\begin{itemize}
\item We developed \snippext\ (snippets of extractions), a system for extracting aspect and opinion pairs, and corresponding sentiments from reviews by fine-tuning a language model with very little labeled training data. \snippext\ is not tied to any language model although we use the state-of-the-art language model BERT for our implementation and experiments as depicted in Figure~\ref{fig:snippext}.
    
\item A novelty of \snippext\ is the clever use of a two-prong
approach to achieve SOTA performance with little labeled training
data: through (1) data augmentation to automatically generate more
labeled training data (\mixda, top-left of Figure~\ref{fig:snippext}), 
and through (2) a semi-supervised learning
technique to leverage the massive amount of unlabeled data in addition
to the (limited amount of) labeled data (\mixmatchnl, right half of Figure~\ref{fig:snippext}).  
The unlabeled data allows
the trained model to better generalize the entire data distribution
and avoid overfitting to the small training set.
%

\item \snippext\ introduces a new data augmentation technique, called \mixda, 
which allows one to only ``partially'' transform a text sequence so that 
the resulting sequence is less likely to be distorted. 
This is done by a non-trivial adaptation of the \mixup\ technique, which we call \mixupnl, 
from computer vision to text (see Section \ref{sec:da}). 
\mixupnl\ uses the convex interpolation technique on the text's language model encoding rather than the original data.
With \mixda, we develop a set of effective data augmentation (\da) operators suitable for opinion mining tasks. 

\item \snippext\ exploits the availability of unlabeled data through a component called \mixmatchnl, which is
a novel adaptation of \mixmatch~\cite{mixmatch} from images to text. \mixmatchnl\ {\em guesses the labels} for unlabeled data and {\em interpolates} 
data with guessed labels and data with known labels.
While the guess and interpolate idea has been carried out in computer vision for training high-accuracy image classifiers, this is the first time the idea is adapted for text.
\mixmatchnl\ leverages \mixupnl(described earlier).
Our data augmentation based on \mixda\ also provides further performance improvement to \mixmatchnl.
\item 
We evaluated the performance of \snippext\ on four Aspect-Based Sentiment Analysis (ABSA)
benchmark datasets.
The highlights of our experimental analysis include: (1)
We achieve new SOTA F1/Macro-F1 scores on all four tasks established by 
\mixda\ and \mixmatchnl\ of \snippext.
(2) Further, a surprising result is that we already achieve the
previous SOTA when given only 1/2 or even 1/3 of the original training data.
\item We also evaluate the practical impact of \snippext\ by applying it to
a large real-world hotel review corpus. Our analysis shows that
\snippext\ is able to extract more fine-grained opinions/customer experiences that are missed by previous methods.
\end{itemize}

\vspace{-1mm}
\noindent
{\bf Outline~}
In Section~\ref{sec:overview}, we overview \snippext\ and its core modules.
We introduce our data augmentation technique \mixda\ in Section \ref{sec:da}.
Section \ref{sec:mixmatch} introduces \mixmatchnl, an adaptation of \mixmatch\ to text.
We show our experiment results in Section \ref{sec:experiment} and \ref{sec:impact}.
Finally, we discuss related work in Sections \ref{sec:related} and conclude in Section \ref{sec:conclusion}.
\vspace{-1mm}

\newcommand{\otag}{\mathsf{O}}
\newcommand{\bastag}{\mathsf{B\text{-}AS}}
\newcommand{\iastag}{\mathsf{I\text{-}AS}}
\newcommand{\boptag}{\mathsf{B\text{-}OP}}
\newcommand{\ioptag}{\mathsf{I\text{-}OP}}

\vspace{-1mm}
\section{Preliminary} \label{sec:overview}

\setlength{\tabcolsep}{2.5pt}
\begin{table*}[ht!]
\small
\vspace{-2mm}
\caption{\small Different tasks in ABSA and \snippext.} \vspace{-3mm}
\label{tab:tasks}
\begin{tabular}{c|c|c|c}
\toprule
\textbf{Tasks}  & \textbf{Task Types} & \textbf{Vocabulary}  & \textbf{Examples Input/Output}\\ \midrule
\begin{tabular}{@{}c@{}}Aspect/Opinion Ext. \\ (similarly for AE in ABSA) \end{tabular}
        & Tagging             & \{B-AS, I-AS, B-OP, I-OP, O\} & 
\setlength{\tabcolsep}{2pt}
\begin{tabular}{@{}cccccccccccccc@{}}
$S = $ & Everybody & was & very & nice & , & but & the & food & was & average & at & best & . \\ 
$\Rightarrow$ & B-AS & O & B-OP & I-OP & O & O & O & B-AS & O & B-OP & I-OP & I-OP & O\end{tabular} \\ \midrule
Aspect Sentiment Cls. & Span Cls. & \{-1, +1, 0\}                &  
$P = \{(1, 1)\}$ (i.e.,``Everybody'') $ \Rightarrow +1$ ; 
$P = \{(8, 8)\}$ (i.e.,``food'') $\Rightarrow 0$ \\
Attribute Cls.        & Span Cls. & Domain-specific attributes & 
$P = \{(1, 1)\}$ $\Rightarrow \mathbf{Staff}$ ; 
$P = \{(8, 8)\}$ $\Rightarrow \mathbf{Food}$ \\
Aspect/Opinion Pairing          & Span Cls. & \{PAIR, NOTPAIR\} &  
$P = \{(1, 1), (3, 4)\} \Rightarrow $ PAIR ; 
$P = \{(8, 8), (3, 4)\} \Rightarrow $ NOTPAIR ($\times$``very nice food'') \\ \bottomrule
\end{tabular}
\vspace{-2mm}
\end{table*}

The goal of \snippext\ is to extract high-quality information from text with small amounts of labeled training data. In this paper, we focus on four main types of extraction tasks, which can be formalized as either tagging or span classification problems.

\vspace{-1mm}
\subsection{Tagging and Span Classification}

\begin{figure}[!ht]
\vspace{-2mm}
    \centering
    \includegraphics[width=0.45\textwidth]{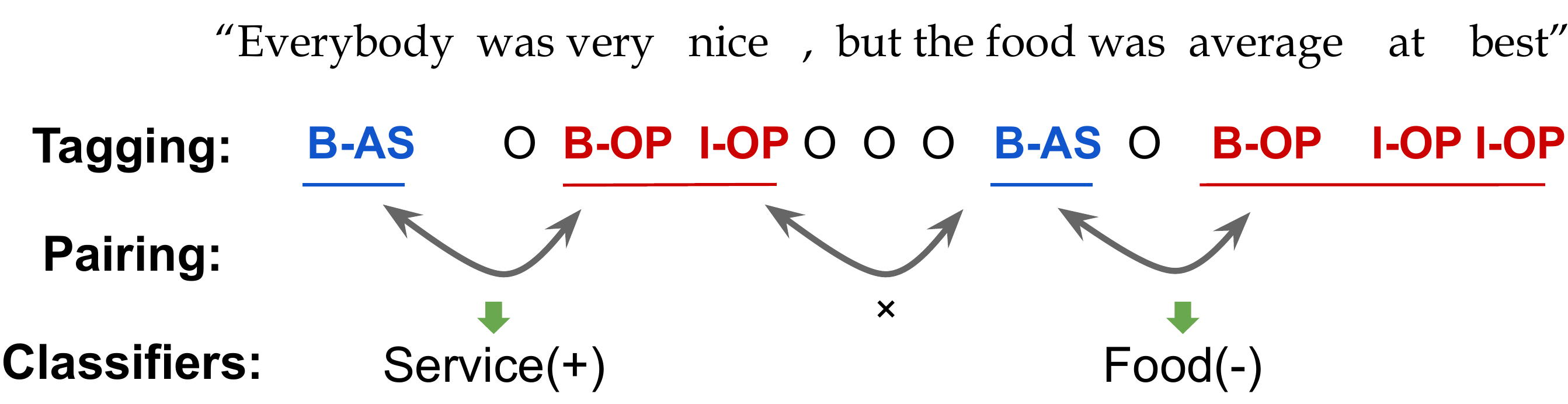}
    \vspace{-2mm}
\caption{\small{The tagging model identifies all aspect (AS) and opinion (OP) spans. 
    Among all candidate pairs of AS-OP spans, the pairing model identifies the correct associations,
    e.g., (``very nice'', ``Everybody'') is correct but not (``very nice'', ``food''). 
    Finally, there are two classifiers decide: (1) which attribute that an extracted pair should be assigned to
    and (2) the sentiment (positive, negative, or neutral) of the opinion.}}
\vspace{-4mm}
    \label{fig:subtasks}
\end{figure}

\noindent
{\bf Types of extraction tasks~}
Figure \ref{fig:extractor} already illustrates the {\em tagging} and {\em sentiment classification} extraction tasks. Implicit in Figure~\ref{fig:extractor} is also the {\em pairing} task that understands which aspect and opinion terms go together.
Figure~\ref{fig:subtasks} makes these tasks explicit, where in addition to tagging, pairing, and sentiment classification, there is
also the {\em attribute classification} task, which determines which attribute a pair of aspect and opinion terms belong to. Attributes are important for downstream applications such as summarization and 
query processing \cite{opinedb, voyageur}.
As we will describe, sentiment classification, pairing, and attribute classification are all instances of the span classification problem.
In what follows, we sometimes refer to an aspect-opinion pair as an opinion.

\begin{definition}\label{def:tagging}
(Tagging) {\em
Let $V$ be a vocabulary of labels.
A tagging model $M$ takes as input a sequence $S = [s_1, \dots, s_n]$ of tokens
and outputs a sequence of labels $M(S) = [l_1, \dots, l_n]$ where each label $l_i \in V$.
}
\end{definition}

Aspect and opinion term extractions are sequence tagging tasks as in ABSA
~\cite{wang2016recursive,wang2017coupled,opinedb,posttrain}, where
$V = \{\bastag, \iastag, \boptag, \ioptag, \\ \otag\}$ using the classic IOB format.
The $\bastag$/$\boptag$ tags indicate that a token is at the beginning of an aspect/opinion term,
the $\iastag$/$\ioptag$ tags indicate that a token is inside an aspect/opinion term and
$\otag$ tags indicate that a token is outside of any aspect/opinion term.

\begin{definition} \label{def:spancls}
(Span Classification) {\em
Let $V$ be a vocabulary of class labels.
A span classifier $M$ takes as input a sequence $S = [s_1, \dots, s_n]$ and 
a set $P$ of spans. Each span $p \in P$ is represented by a pair of indices 
$p = (a, b)$ where $1 \leq a \leq b \leq n$ indicating the start/end positions of the span.
The classifier outputs a class label $M(S, P) \in V$.
}
\end{definition}

Both Aspect Sentiment Classification (ASC) \cite{posttrain,auxiliarysentence} and the aspect-opinion
pairing task can be formulated as span classification tasks \cite{opinedb}.
For ASC, the span set $P$ contains a single span which is the targeted aspect term.
The vocabulary $V = \{+1, 0, -1\}$ indicates positive, neutral, or negative sentiments.
For pairing, $P$ contains two spans: an aspect term and an opinion term.
The vocabulary $V = \{\mathsf{PAIR}, \mathsf{NOTPAIR}\}$ indicates whether
the two spans in $P$ are correct pairs to be extracted or not.
Attribute classification can be captured similarly.
Table \ref{tab:tasks} summarizes the set of tasks considered in ABSA and \snippext.

\subsection{Fine-tuning Pre-trained Language Models}

Figure~\ref{fig:snippext} shows the basic model architecture in \snippext, where it makes use of a pre-trained language model (LM).

Pre-trained LMs such as BERT \cite{bert}, GPT-2 \cite{gpt2}, and XLNet \cite{xlnet} have
demonstrated good performance in a wide range of NLP tasks. 
In our implementation, we use the popular BERT language model although our proposed techniques 
(detailed in Sections \ref{sec:da} and \ref{sec:mixmatch}) 
are independent of the choice of LMs. 
We optimize BERT by first fine-tuning it with a domain-specific text corpus
then fine-tune the resulting model for the different subtasks. 
This has been shown to be a strong baseline for various NLP tasks \cite{scibert,biobert,posttrain}
including ABSA. 



\smallskip
\noindent
\textbf{Fine-tuning LMs for specific subtasks. }
Pre-trained LMs can be fine-tuned
to a specific task through a task-specific labeled training set as follows:

\begin{enumerate}
\item Add task-specific layers (e.g., a simple fully connected layer
for classification) after the final layer of the LM;
\item Initialize the modified network with parameters from the pre-trained model;
\item Train the modified network on the task-specific labeled data.
\end{enumerate}

\begin{figure}[!ht]
\vspace{-1mm}
    \centering
    \includegraphics[width=0.47\textwidth]{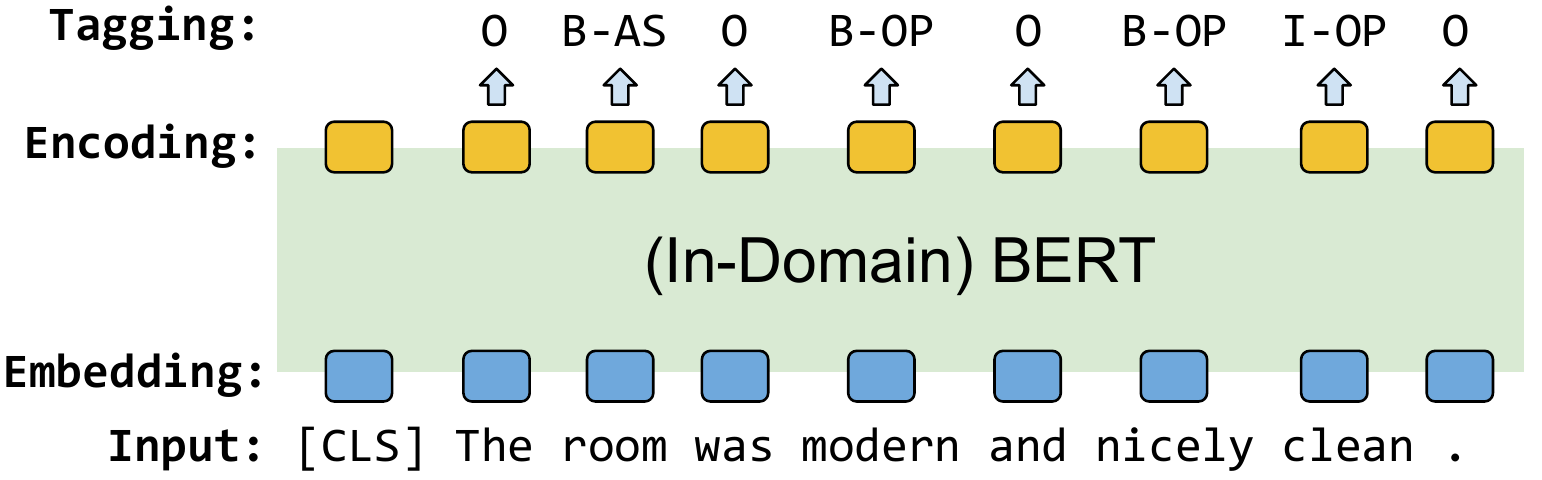}
\vspace{-2mm}
    \caption{\small{Fine-tuning BERT for aspect/opinion term extraction.}}
\vspace{-3mm}
    \label{fig:bert}
\end{figure}

We fine-tune BERT to obtain our
tagging and span classification models.
For both tagging and span classification, 
the task-specific layers consist of only one fully connected layer followed 
by a softmax output layer. 
The training data also need to be encoded 
into BERT's  
input format. We largely follow the fine-tuning approach described in \cite{bert,posttrain} and 
Figure \ref{fig:bert} shows an example of the model architecture for
tagging aspect/opinion terms.  We 
describe more details
in Section \ref{sec:experiment}.

As mentioned earlier, our proposed techniques
are independent of the choice of LMs and task-specific layers. We use the basic 12-layer uncased BERT and one fully connected task-specific layer in this paper but one can also use higher-quality models with deeper LMs (e.g., a larger BERT
or XLNet) or adopt more complex task-specific layers (e.g., LSTM and CRF) to further improve the results obtained. 

\smallskip
\noindent
\textbf{Challenges in optimizing LMs. } It has been shown that fine-tuning BERT for specific tasks
achieves good results, often outperforming previous neural network models
for multiple tasks of our interest \cite{posttrain,sun2019utilizing}.
However, like in many other deep learning approaches,
to achieve good results on fine-tuning for specific tasks requires a fair amount of quality labeled training data (e.g., 3,841 labeled sentences were used for aspect term extraction for restaurants~\cite{pontiki2014semeval, pontiki2015semeval}) and
creating such datasets with desired quality is often expensive. 

\snippext\ overcomes the requirement of having a large quality labeled training set by addressing the following two questions:
\begin{itemize}
\item[(1)] Can we 
make the best of a small set of labeled training data by generating high-quality training examples from it? 
\item[(2)] Can we leverage BOTH labeled and unlabeled data for fine-tuning the in-domain LM for specific tasks and obtain better results?
\end{itemize}
We address these two questions in Sections \ref{sec:da} and \ref{sec:mixmatch} respectively.

\newcommand{\tfidf}{\ensuremath{\mathsf{TFIDF}}}
\newcommand{\similarity}{\ensuremath{\mathsf{sim}}}
\newcommand{\bert}{\ensuremath{\mathsf{BERT}}}
\newcommand{\aug}{\ensuremath{\text{aug}}}

\vspace{-1mm}
\section{MixDA: augment and interpolate} \label{sec:da}
Data augmentation (\da) is a technique to 
automatically 
increase the size of the training data without
using human annotators. DA is shown to be effective for certain tasks in computer vision and NLP.
In computer vision, 
labeled images that are augmented through simple operators such as rotate, crop, pad, flip are shown to be 
effective for training
deep neural networks \cite{perez2017effectiveness,autoaugment}.
In NLP, sentences that are augmented by 
replacing tokens with their corresponding synonyms are shown to be effective for training sentence classifiers~\cite{eda}.
Intuitively, such augmented data
allow the trained model to learn properties 
that remain invariant in the data (e.g., the meaning of a sentence remains unchanged if a token is replaced with its synonym).
However, in NLP tasks the use of DA is still limited, as synonyms of a word are very limited and other operators can distort the meaning of the augmented sentence. Motivated by above issues and inspired by the ideas of data augmentation and \mixup~\cite{mixup}, we introduce the \mixda\ technique that generates augmented data through (1) carefully augmenting the set of labeled sentences through a set of data augmentation operators 
suitable for 
tagging and span-classification, and (2) performing a convex interpolation on the augmented data with the original data to further reduce the noise that may occur in the augmented data.
\mixda\ uses the resulting interpolation as the training signal.

\vspace{-1mm}
\subsection{Data Augmentation Operators}
\label{sec:regularda}
The typical data augmentation operators that have been proposed for text~\cite{eda,xie2019unsupervised} 
include: token replacement 
(replaces a token with a new one in the example); 
token insertion 
(inserts a token into the example); 
token deletion 
(removes a token from the example); 
token swap 
(swaps two tokens in the example); 
and back translation 
(translates the example into a different language and back, e.g., EN $\rightarrow$ FR $\rightarrow$ EN).

Although these operators were shown to be effective in augmenting training data for sentence classification, 
a naive application of these operators can be problematic for the tagging or span classification tasks as the following example illustrates.

\begin{center}
\small
\begin{tabular}{ccccccc}
The & \textbf{food} & was & \textbf{average} & \textbf{at} &\textbf{best} &.\\
O&\textbf{B-AS} & O&\textbf{B-OP}&\textbf{I-OP}&\textbf{I-OP}&O\\
\end{tabular} 
\end{center} 
A naive application of swap or delete may leave
the sequence with an inconsistent state of tags (e.g., if ``average'' was removed, I-OP is no longer preceded by B-OP).
Even worse, replace or insert  can change the meaning of tokens and also make the original tags invalid (e.g., by replacing ``at'' with ``and'', the correct tags should be ``average (B-OP) and(O) best(B-OP)''). 
Additionally, back translation changes
the sentence structure and tags
are lost during the translation.

The above example suggests that \da\ operators must be carefully applied. Towards this, we distinguish two types of tokens. We call 
the 
consecutive tokens with non-``$\otag$'' tags (or consecutive tokens represented by a pair of 
indices in span classification tasks), the {\em target spans}.  The tokens within target spans are {\em target tokens}
and the tokens with ``$\otag$'' tags, are the {\em non-target tokens}. 
To guarantee the correctness 
of the tagging sequence, we apply $\da$ operators over target spans and non-target tokens.
Specifically, we consider only  4 token-level operators similar to what was described earlier (\textbf{TR} (replace), \textbf{INS} (insert), \textbf{DEL} (delete), and \textbf{SW} (swap)) but 
apply them only on non-target tokens.
We also introduced a new span-level operator  (\textbf{SPR} for span-level replacement), 
which augments the input sequences by replacing a target span 
with a new span of the same type. Table~\ref{tab:da} summarizes the set of $\da$ operators in \snippext.

\setlength{\tabcolsep}{4pt}
\begin{table}[!h]
\small
\vspace{-4mm}
\caption{\small \da\ operators of \snippext. TR, INS, DEL, SW are modified from prior operators. SPR is a new span-level operator.} \vspace{-3mm}
\vspace{-1mm}
\label{tab:da}
\begin{tabular}{c|c} \toprule
\textbf{Operator}  &{\textbf{Description}}\\ \midrule
\textbf{TR} & Replace \textbf{non-target token} with a new token. \\ 
\textbf{INS} & Insert before or after a \textbf{non-target token} with a new token. \\
\textbf{DEL}       & Delete a \textbf{non-target token}. \\ 
\textbf{SW}         & Swap two \textbf{non-target tokens}. \\\midrule
\textbf{SPR}     & Replace a \textbf{target span} with a new span.\\\bottomrule
\end{tabular}
\vspace{-4mm}
\end{table}

To apply a $\da$ operator, 
we first sample a token (or span) from the original example.
If the operator is \textbf{INS}, \textbf{TR}, or \textbf{SPR}, then we also need to perform a 
post-sampling step to determine a new token (or span) to insert or replace the original one. There are two strategies for sampling (and one more for post sampling):

\begin{itemize}
\item Uniform sampling: picks a token or span from the sequence with equal probability. This is a commonly used strategy as in~\cite{eda}.

\item Importance-based sampling: picks a token or span based on probability proportional to the importance of the token/span, which is measured by the token's TF-IDF~\cite{xie2019unsupervised} or the span's frequency.
\item Semantic Similarity (post-sampling only): picks a token or span with probability proportional to its semantic similarity with the original token/span. Here, we measure the semantic similarity by the cosine similarity over token's or span's embeddings\footnote{For token, we use  Word2Vec \cite{word2vec} embeddings; for spans, we use the BERT encoding.}.
\end{itemize}

For {\bf INS}/{\bf TR}/{\bf SPR}, the post-sampling step will pick a similar token (resp. span) to insert or to replace the token (resp. span) that was picked in the pre-sampling step.
We explored different combinations of pre-sampling and post-sampling strategies and report the most effective strategies in Section~\ref{sec:experiment}. 

\vspace{-1mm}
\subsection{Interpolate} \label{sec:mixda}
Although the $\da$ operators are designed
to minimize distortion to the original sentence
and its labels, the operators can still generate examples that are ``wrong'' with regard to certain labels. 
As we found in Section \ref{sec:exp:da},
these wrong labels can make the \da\ operator less effective or 
even hurt the resulting model's performance.

\vspace{-1mm}
\begin{example}
Suppose the task is to classify the aspect sentiment
of ``Everybody'':
\begin{center}
\underline{Everybody} (+1) was very nice ...
\end{center}
The $\da$ operators can still generate
examples that are wrong with regard to the labels.
For example,
\textbf{TR} may replace ``nice'' with a negative/neutral word (e.g., ``poor'', ``okay'')
and hence the sentiment label is no longer +1.
Similarly, \textbf{DEL} may drop ``nice'', \textbf{INS} 
may insert ``sometimes'' after ``was'', or \textbf{SPR} can replace ``Everybody'' with ``Nobody'' so that the sentiment label would now be wrong.
\end{example}
\vspace{-1mm}

To reduce the noise that may be introduced
by the augmented data,
we propose a novel technique called \mixda\ that performs
a convex interpolation on the augmented data with the original data and uses the interpolated
data as training data instead.
Intuitively, the interpolated result is
an intermediary between an example and an augmented example.
By taking this ``mixed'' example, which is only``partially augmented'' and ``closer'' to the original example than the augmented example, the resulting training data are likely to contain less distortion.

\smallskip
\noindent \textbf{\textbf{MixUp}$\mathbf{^{NL}}$. }
Let $x_1$ and $x_2$ be two text sequences and $y_1$ and $y_2$ as their one-hot label vectors\footnote{For tagging task, $y_1$ and $y_2$
are sequences of one-hot vectors.} respectively.
We assume that both sequences are padded into the same length.
We first create two new sequences $x'$ and $y'$ as follows:
\vspace{-1mm}
\begin{eqnarray}
\bert(x') &=& \lambda \cdot \bert(x_1) + (1-\lambda) \cdot \bert(x_2) \label{interpolation:x}\\
y' &=& \lambda \cdot y_1 + (1-\lambda) \cdot y_2 \label{interpolation:y}
\vspace{-1mm}
\end{eqnarray}
where $\bert(x_1)$ and $\bert(x_2)$ are the BERT encoding for $x_1$ and $x_2$, $y_1$ are original labels of $x_1$, $y_2$ labels are generated directly from $y_1$ when performing DA operators, \zhengjie{expand a bit according to our rebuttal} and
$\lambda \in [0, 1]$ is a random variable sampled from a symmetric 
Beta distribution $\text{Beta}(\alpha, \alpha)$ for a hyper-parameter $\alpha$.
Note that we do not actually generate the interpolated sequence $x'$ but instead,
only use the interpolated encoding $\bert(x')$ to carry out the computation
in the task-specific layers of the neural network.
Recall that the task-specific layers for tagging take as input the entire $\bert(x')$
while the span-classification tasks only require the encoding of the aggregated ``$\mathsf{[CLS]}$'' token.


\begin{figure}[!bht]
\vspace*{-3mm}
    \centering
    \includegraphics[width=0.43\textwidth]{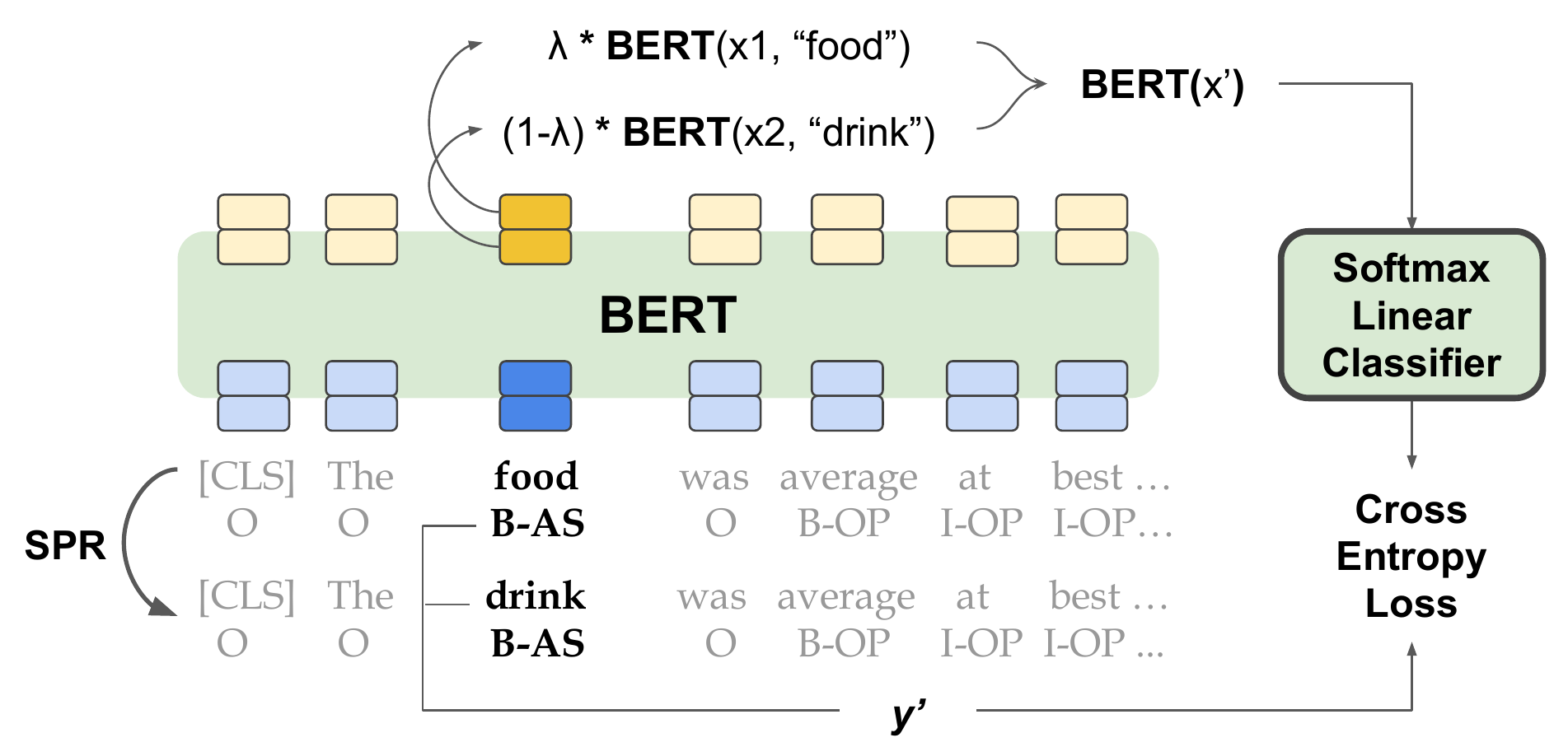}
\vspace*{-4mm}
    \caption{\small MixDA augments by interpolating over the BERT encoding.}
\vspace*{-3mm}
    \label{fig:mixda}
\end{figure}

Figure~\ref{fig:mixda} illustrates how we train a model using the results of $\mixda$. 
Given an example $(x, y)$, $\mixda$ trains a model through three main steps:
\vspace{-1mm}
\begin{itemize}
\item \textbf{Data Augmentation}: a $\da$ operator (see Section~\ref{sec:regularda}) is applied to obtain $(x_\aug, y_\aug)$. 
\item \textbf{Interpolation}: perform the $\mixupnl$ interpolation on the 
pair of input $(x, y)$ and $(x_\aug, y_\aug)$ to obtain $(\bert(x'), y')$. 
The resulting $\bert(x')$ corresponds to the encoding of a sequence
``in between'' the original sequence $x$ and the fully augmented sequence $x_\aug$.
\item \textbf{Back Propagation}: feed the interpolated encoding $\bert(x')$
to the remaining layers, compute the loss over $y'$, and back propagate to update the model.

\item \textbf{Data Augmentation Optimization}:  
Since \da\ operators may change the sequence length, for tagging,
$\mixda$ also carefully aligns the
label sequence $y_\aug$  with $y$. This is done by padding tokens
to both $x$ and $x_\aug$ when the inserting/deleting/replacing of tokens/spans 
creates  misalignments in the two sequences.
When the two sequences are perfectly aligned, Equation~\ref{interpolation:y}
simplifies to $y'=y_1$. 
\end{itemize}
\vspace{-1mm}
Intuitively, by the interpolation, 
\mixda\ allows an input sequence to be augmented by a \da\ operator partially
(by a fraction of $1 - \lambda$) to effectively reduce the distortion produced by the 
original operator.
Our experiment results in Section \ref{sec:exp:da} confirms that
applying \da\ operators with \mixda\ is almost always beneficial (in 34/36 cases) and 
can result in up to 2\% performance improvement in aspect sentiment classification.

\noindent
{\bf Discussion.~}The interpolation step of
$\mixda$ is largely inspired by the $\mixup$ operator \cite{mixup,verma2018manifold} in computer vision, which 
has been shown to be 
a very effective regularization technique for learning better image representations. \mixup\
produces new training examples by combining two existing examples through their convex interpolation. With
the interpolated examples as training data,
the trained model can now make predictions that
are ``smooth'' in between the two examples. For example, in a binary classification of 
cat and dog images, the model would learn that (1) the ``combination'' of two cats (or dogs) should be
classified as cat (or dog); (2) something ``in between'' a cat and a dog should be given less confidence, i.e., the model should predict both classes with similar probability.

Unlike images, however, text sequences are
not continuous and have different lengths.
Thus, we cannot apply convex interpolation directly 
over the sequences. In \snippext, 
we compute the language model's encoding of the two sequences 
and interpolate the encoded sequences instead. 
A similar idea was considered in computer vision \cite{verma2018manifold}
and was shown to be more effective than directly interpolating the inputs.
Furthermore, in contrast to image transformations that generate a continuous range of training examples in the vicinity of the original image, 
traditional text \da\ operators only generate a limited finite set of examples. 
\mixda\ increases the coverage of \da\ operators by
generating varying degrees of partially augmented training examples.
Note that \mixup\ has been applied in NLP in a setting \cite{guo2019augmenting} with sentence classification and CNN/RNN-based models.
To the best of our knowledge, \mixda\ is the first to apply \mixup\ on text with a pre-trained LM and data augmentation.
\newcommand{\sharpen}{\ensuremath{\mathsf{Sharpen}}}
\newcommand{\loss}{\ensuremath{\mathsf{Loss}}}
\newcommand{\model}{\ensuremath{\mathsf{Model}}}
\newcommand{\enc}{\ensuremath{\mathsf{enc}}}
\newcommand{\virtual}{\ensuremath{\mathsf{v}}}

\vspace{-2mm}
\section{Semi-Supervised Learning with MixMatch$^{NL}$} \label{sec:mixmatch}
Semi-supervised learning (SSL) is the learning paradigm \cite{sslsurvey} where models
learn from a small amount of labeled data and a large amount of
\emph{unlabeled data}. 
We propose \mixmatchnl, a novel SSL framework for NLP
based on an adaptation of  \mixmatch\ \cite{mixmatch}, which is a recently proposed technique in computer vision
for training high-accuracy image classifiers with limited amount of labeled images.

\smallskip
\noindent
\textbf{Overview. } 
As shown in Figure \ref{fig:snippext} earlier, \mixmatchnl\ leverages the massive amount of unlabeled data by 
\emph{label guessing} and \emph{interpolation}.
For each unlabeled example, \mixmatchnl\ produces a ``soft'' (i.e., continuous) 
label (i.e., the guessed label) 
predicted by the current model state. The guessed labeled example can now be used as training data. However, it can be noisy due to the current model's quality.
Thus, like in \mixmatch\ which does not use the guessed labeled example directly,
we interpolate this guessed labeled example with a labeled one and use the interpolated result for training instead.
However, unlike \mixmatch\ which interpolates two images, \mixmatchnl\ interpolates two text sequences by applying the \mixupnl\ idea again that was previously described in \mixda.
Instead of interpolating the guessed labeled example with the labeled example directly,
we interpolate the two sequences' encoded representation that we obtain from
a language model such as BERT.
The interpolated sequences and labels are then fed into the remaining layers and we
compute the loss and back-propagate to update the network's parameters.

\mixmatchnl\ also benefits from the integration with \mixda.
As we will show in Section \ref{sec:exp:mixmatch},
replacing the normal \da\ operators with \mixda\ allows 
\mixmatchnl\ to achieve a performance improvement of up to 1.8\% in our experiment
results with opinion mining tasks. 
Combining \mixda\ and by leveraging the unlabeled data,
\mixmatchnl\ effectively reduces the requirement of labeled data by 50\% or more
to achieve previous SOTA results.  

We now describe each component.
\mixmatchnl\ takes as input a batch $B$ of labeled examples
$X = \{(x_b, y_b)\}_{1\leq b \leq B}$ and a batch of unlabeled examples
$U = \{u_b\}_{1 \leq b \leq B}$. 
Each $x_b$ and $u_b$ is a text sequence and $y_b$ is an one-hot vector
(or a sequence of one-hot vectors for tagging tasks) representing the label(s) of $x_b$.
We assume that sequences in $X$ and $U$ are already padded into the same length.
Like in MixMatch, \mixmatchnl\ augments and mixes the two batches and 
then uses the mixed batches as training signal in each training iteration.
This is done as follows.

\smallskip
\noindent
\textbf{Data Augmentation. } Both $X$ and $U$ are first augmented with the $\da$ operators.
So every $(x, y) \in X$, is augmented
into a new example 
$(\hat{x}, \hat{y})$. We denote by $\hat{X}$ the augmented labeled examples.
Similarly, each unlabeled example $u_b \in U$ is augmented into $k$ 
examples $\{\hat{u}_{b,1}, \dots, \hat{u}_{b,k}\}$ for a hyper-parameter $k$.

\smallskip
\noindent
\textbf{Label Guessing. } Next, we guess the label for each unlabeled example 
in $U$. Each element of a guessed label of $u_b \in U$ is a probability distribution over the label vocabulary
computed as the average of the model's current prediction on the $k$ augmented examples of $u_b$.
Formally, the guessed label $\bar{q}_b$ is computed as
\vspace{-2mm}
$$  \bar{q}_b = \dfrac{1}{k} \sum_{j=1}^k \model(\hat{u}_{b,j}) \vspace{-2mm}$$
where $\model(\hat{u}_{b,j})$ is the label distribution output of the model on the example $\hat{u}_{b,j}$
based on the current model state. 

In addition, to make the guessed distribution closer to an one-hot distribution,
\mixmatchnl\ further reduces the entropy
of $\bar{q}_b$ by computing $q_b = \sharpen(\bar{q}_b)$.
$\sharpen$ is an element-wise sharpening function where
for each guessed distribution $p$ in $q_b$:
\vspace{-2mm}
$$ \sharpen(p)_i := p_i^{1/T} \bigg/ \sum_{j=1}^{v} p_j^{1/T} \vspace{-2mm}$$ 
$v$ is the vocabulary size and $T$ is a hyper-parameter in the range $[0, 1]$.
Intuitively, by averaging and sharpening the multiple ``guesses'' on the augmented examples,
the guessed label $q_b$ becomes more reliable as long as most guesses are correct.
The design choices in this step largely follow the original MixMatch.
To gain further performance improvement,
we generate each $\hat{u}_{b,j}$ with \mixda\ instead of regular \da.
We set $k=2$ for the number of guesses.

\smallskip
\noindent
\textbf{Mixing Up. } The original MixMatch requires interpolating
the augmented labeled batch $\hat{X} = \{(\hat{x}_b, y_b)\}_{1\leq b \leq B}$
and the unlabeled batch with guessed labels $\hat{U} = \{\hat{u}_{b,j}, q_b)\}_{1\leq b \leq B, 1 \leq j \leq k}$,
but it is not trivial how to interpolate text data.
We again use \mixupnl's idea of interpolating the LM's output.
In addition, we also apply \mixda\ in this step to 
improve the performance of the \da\ operators.
Formally, we
\begin{itemize}
\item[(1)] Compute the LM encoding $\enc(\cdot)$ of $X$, $\hat{X}$, and $\hat{U}$ where
\begin{align*}
\enc(X) &= \{(\bert(x_b), y_b) \}_{1 \leq b \leq B}, \\
\enc(\hat{X}) &= \{(\bert(\hat{x}_b), y_b) \}_{1 \leq b \leq B}, \\
\enc(\hat{U}) &= \{(\bert(\hat{u}_{b,j}), q_b) \}_{1 \leq b \leq B, 1 \leq j \leq k}.
\end{align*}

\item[(2)] Sample $\lambda_1 \sim \text{Beta}(\alpha_{\text{aug}}, \alpha_{\text{aug}})$
$\lambda_2 \sim \text{Beta}(\alpha_{\text{mix}}, \alpha_{\text{mix}})$ for
two given hyper-parameters $\alpha_{\text{aug}}$ and $\alpha_{\text{mix}}$.
Here $\lambda_1$ is the interpolation parameter for $\mixda$ and
$\lambda_2$ is the one for mixing labeled data with unlabeled data.
We set $\lambda_2 \leftarrow \max\{\lambda_2, 1 - \lambda_2\}$ to ensure
that the interpolation is closer to the original batch.

\item[(3)] Perform \mixda\ between $X$ and $\hat{X}$. We use the notation $^\virtual$ to represent \emph{virtual} examples not generated but whose LM encodings are obtained by interpolation. Let $\hat{X}^{\virtual}$ be the \mixda\
interpolation of $X$ and $\hat{X}$, and $\enc(\hat{X}^{\virtual})$ be its LM encoding. We have
$$ \enc(\hat{X}^{\virtual}) = \lambda_1 \cdot \enc(X) + (1 - \lambda_1) \cdot \enc(\hat{X}) . $$

\item[(4)] Shuffle the union of the \mixda\ output $\enc(\hat{X}^{\virtual})$
and the LM encoding $\hat{U}$, then mix with $\enc(\hat{X}^{\virtual})$ and $\enc(\hat{U})$.
Let $X^{\virtual}$ and $U^{\virtual}$ be the virtual interpolated labeled and unlabeled batch
and their LM encodings be $\enc(X^{\virtual})$ and $\enc(U^{\virtual})$ respectively.
We compute:
\begin{align*}
W &= \text{Shuffle}(\text{ConCat}(\enc(\hat{X}^{\virtual}), \enc(\hat{U})) ) , \\
 \enc(X^{\virtual}) &= \lambda_2 \cdot \enc(\hat{X}^{\virtual}) + (1 - \lambda_2) \cdot W_{[1 \dots B]} , \\
 \enc(U^{\virtual}) &= \lambda_2 \cdot \enc(\hat{U}) + (1 - \lambda_2) \cdot W_{[B+1 \dots (k+1)B]} .
\end{align*}
\end{itemize}
In essence, we ``mix'' $\hat{X}$ with the first $|B|$ examples of $W$
and $\hat{U}$ with the rest. 
The resulting $\enc(X^{\virtual})$ and $\enc(U^{\virtual})$ 
are batches of pairs $\{(\bert(x^{\virtual}_b), y^{\virtual}_b)\}_{1 \leq b \leq B}$ and
$\{(\bert(u^{\virtual}_{b, j}), q^{\virtual}_b)\}_{1 \leq b \leq B, 1 \leq j \leq k}$
where each $\bert(x^{\virtual}_b)$ (and $\bert(u^{\virtual}_{b,j})$) is an interpolation of two BERT representations.
The interpolated text sequences, $x^{\virtual}_b$ and $u^{\virtual}_{b,j}$, are not actually generated.

Note that $\enc(X^{\virtual})$ and $\enc(U^{\virtual})$ contain
interpolations of (1) labeled examples, (2) unlabeled examples, and (3) pairs of labeled and unlabeled examples.
Like in the supervised setting,
the interpolations encourage the model to make smooth transitions ``between''
examples. In the presence of unlabeled data,
such regularization is imposed not only between pairs of labeled data
but also unlabeled data and pairs of label/unlabeled data.

The two batches $\enc(X^{\virtual})$ and $\enc(U^{\virtual})$ are then fed into the remaining layers
of the neural network to compute the loss and back-propagate to update
the network's parameters.




\smallskip
\noindent
\textbf{Loss Function. } 
Similar to MixMatch, \mixmatchnl\ also adjusts the loss function to take into account
the predictions made on the unlabeled data.
The loss function is the sum of two terms: 
(1) a cross-entropy loss between the predicted label distribution with the groundtruth label and 
(2) a Brier score (L2 loss) for the unlabeled data which is less sensitive to
the wrongly guessed labels. 
Let $\model(x)$ be the model's predicted probability distributions
on BERT's output $\bert(x)$. 
Note that $x$ might be an interpolated sequence in $X^{\virtual}$ or $U^{\virtual}$
without being actually generated.
The loss function is
$\loss(\enc(X^{\virtual}), \enc(U^{\virtual})) = \loss_X + \lambda_U \loss_U$ where
\vspace{-1mm}
\begin{align*}
 \loss_X &= \dfrac{1}{|X^{\virtual}|} \sum_{\bert(x),y \in \enc(X^{\virtual})} \text{CrossEntropy}(y, \model(x)), \\
 \loss_U &= \dfrac{1}{|\mathsf{vocab}| \cdot |U^{\virtual}|} \sum_{\bert(u),q \in \enc(U^{\virtual})} \big\|q - \model(u)\big\|_2 .
\vspace{-3mm}
\end{align*}
The value $B$ is the batch size, $|\mathsf{vocab}|$ is the size of the label vocabulary and 
$\lambda_U$ is the hyper-parameter
controlling the weight of unlabeled data at training.
Intuitively, this loss function encourages the model to make prediction consistent to the guessed labels
in addition to correctly classifying the labeled examples.

\newcommand{\bertfd}{\ensuremath{\mathsf{BERT\text{-}FD}}}
\newcommand{\bertpt}{\ensuremath{\mathsf{BERT\text{-}PT}}}
\newcommand{\bertptm}{\ensuremath{\mathsf{BERT\text{-}PT^-}}}

\vspace{-2mm}
\section{Experiments on ABSA tasks} \label{sec:experiment}

Here, we evaluate the effectiveness of \mixda\ and \mixmatchnl\
by applying them on two ABSA tasks: Aspect Extraction (AE) and Aspect Sentiment Classification (ASC).
On four ABSA benchmark datasets, 
\mixda\ and \mixmatchnl\ achieve previous SOTA results 
(within 1\% difference or better) using only 50\% or less of the training data 
and outperforms SOTA (by up to 3.55\%) when full data is in use.
Additionally, we found that although $\da$ operators can 
result in different performance on different datasets/tasks,
applying them with $\mixda$ is generally beneficial.
\mixmatchnl\ further improves the performance when unlabeled data are taken into account
especially when given even fewer labels ($\leq$ 500).

\vspace{-2mm}
\subsection{Experimental Settings}\label{sec:exp:data}

\setlength{\tabcolsep}{3pt}

\begin{table}[!b]
\small
    \centering
\vspace*{-4mm}
    \caption{\small Some statistics for the benchmark ABSA datasets. S: number of sentences; 
    A: number of aspects; P, N, and Ne: number of positive, negative and neutral polarities.} 
\vspace*{-3mm}
    \label{tab:absa-data}
    \resizebox{\linewidth}{!}{
    \begin{tabular}{c|c|c|c}\toprule
    & \textbf{AE} & \textbf{ASC} & \textbf{LM Fine-tuning}\\ \midrule
\textbf{Restaurant} & SemEval16 Task5 & SemEval14 Task4 &Yelp \\ \midrule
Train & 2000 S / 1743 A & 2164 P / 805 N / 633 Ne & 2M sents \\
Test & 676 S / 622 A & 728 P / 196 N / 196 Ne & - \\ 
Unlabeled & 50,008 S & 35,554 S & -\\ \midrule
\textbf{Laptop} & SemEval14 Task4 & SemEval4 Task4  & Amazon\\ \midrule
Train & 3045 S / 2358 A & 987 P / 866 N / 460 Ne & 1.17M sents \\
Test & 800 S / 654 A & 341 P / 128 N / 169 Ne & - \\ 
Unlabeled & 30,450 S & 26,688 S & -\\ \bottomrule
    \end{tabular}}
\end{table}

\noindent \textbf{Datasets and Evaluation Metrics.}
We consider 4 SemEval ABSA datasets \cite{pontiki2014semeval, absa} from two domains (restaurant and laptop) over the two tasks (AE and ASC).
Table~\ref{tab:absa-data} summarizes the 4 datasets.
We split the datasets into training/validation sets following the settings in \cite{posttrain}, 
where 150 examples from the training dataset are held for validation for all tasks.
For each domain, we create an in-domain BERT model by fine-tuning on raw review text.
We use 1.17 million sentences from Amazon reviews \cite{amazondata} for the laptop domain and 
2 million sentences from Yelp Dataset reviews \cite{yelpdataset} for the restaurant domain. 
These corpora are also used for sampling unlabeled data for \mixmatchnl\ 
and training Word2Vec models when needed. 
We use a baseline AE model to generate aspects for the ASC unlabeled sentences.
We use F1 as the evaluation metric for the two AE tasks and Macro-F1 (MF1) for the ASC tasks.

\smallskip
\noindent \textbf{Varying Number of Training Examples.} 
We evaluate the performance of 
different methods when the size of training data is varied.
Specifically, for each dataset, we vary the number of training examples from 250, 500, 750, to 1000. We create 3 uniformly sampled subsets of each size and 
run the method 5 times on each sample resulting in 15 runs.
For a fair comparison, we also run the method 15 times on all the training data (full).
We report the average results (F1 or MF1) on the test set of the 15 runs. 




\smallskip
\noindent
\textbf{Implementation Details. } 
All evaluated models are based on the 12-layer uncased BERT \cite{bert} model\footnote{Our implementation is based on HuggingFace Transformers \url{https://huggingface.co/transformers/}.
We open-sourced our code: \url{https://github.com/rit-git/Snippext_public}.}.
We use HuggingFace's default setting for the in-domain fine-tuning of BERT.
In all our experiments,
we fix the learning rate to be 5e-5, batch size to 32, and max sequence length to 64.
The training process runs a fixed number of epochs depending on the dataset size
and returns the checkpoint with the best performance evaluated on the dev-set.

\begin{figure*}[!htb]
\minipage{1.0\textwidth}
\begin{center}
\minipage{0.55\textwidth}
  \includegraphics[width=\linewidth]{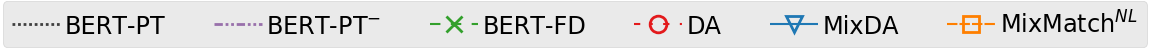}
\endminipage
\end{center}
\minipage{0.24\textwidth}
  \includegraphics[width=\linewidth]{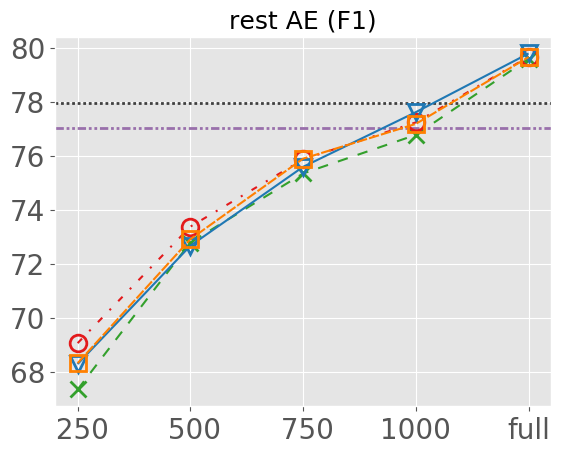}
\endminipage\hfill
\minipage{0.24\textwidth}
  \includegraphics[width=\linewidth]{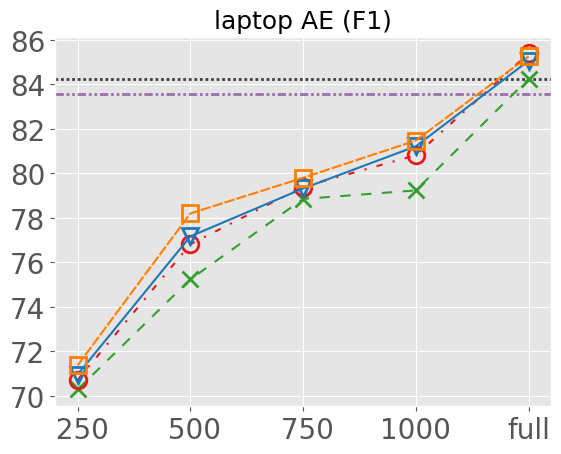}
\endminipage\hfill
\minipage{0.24\textwidth}%
  \includegraphics[width=\linewidth]{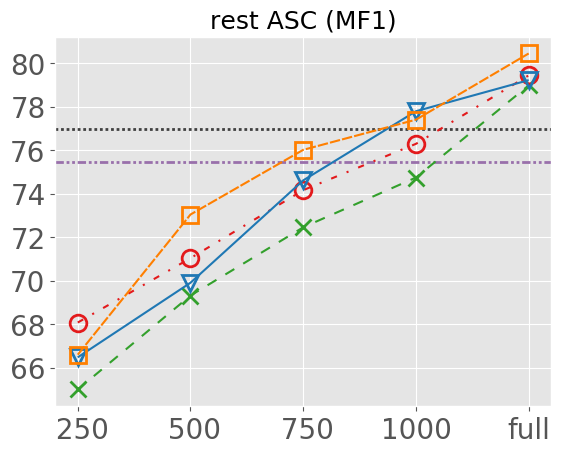}
\endminipage\hfill
\minipage{0.24\textwidth}%
  \includegraphics[width=\linewidth]{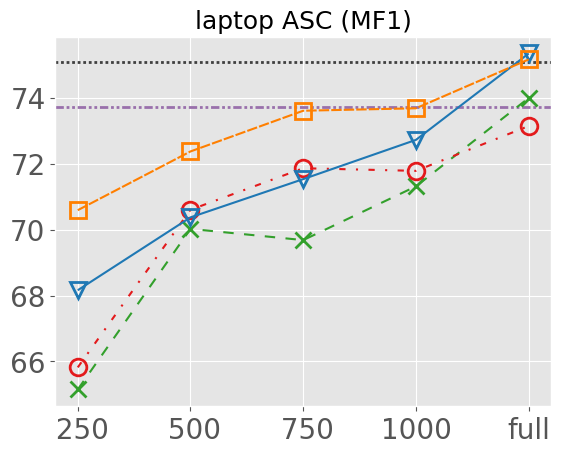}
\endminipage
\endminipage
\vspace{-3mm}
\caption{\small Performance of \da, \mixda, and \mixmatchnl\ on 4 ABSA datasets at different training set sizes. }\label{fig:expmain}
\vspace{-4mm}
\end{figure*}

\smallskip
\noindent
\textbf{Evaluated Methods. } 
In previous work, methods based on fine-tuning pre-trained LMs achieve SOTA results in ABSA tasks. 
We compare \mixda\ and \mixmatchnl\ with these methods as baselines.
\begin{itemize}
\item \textbf{BERT-PT}~\cite{posttrain} (SOTA): \bertpt\ achieves state-of-the-art 
performance on multiple ABSA tasks. Note that in addition to post-training in-domain BERT, \bertpt\ largely leverages an extra labeled reading comprehension dataset. 
\item \textbf{BERT-PT$^-$}~\cite{posttrain}: Unlike \bertpt, \bertptm fine-tunes on the specific tasks without the labeled reading comprehension dataset. 
\item \textbf{BERT-FD}: This is our implementation of fine-tuning in-domain BERT
on specific tasks. \bertfd\ is similar to \bertptm\ except that it leverages
a more recent BERT implementation.
\item \textbf{DA (Section~\ref{sec:regularda})}: \da\ extends \bertfd\ by augmenting the training set through applying a \emph{single} data augmentation operator.
\item \textbf{MixDA (Section~\ref{sec:mixda})}: \mixda\ optimizes \da\ by interpolating the augmented example with the original example.
\item \textbf{MixMatch\textsuperscript{NL} (Section~\ref{sec:mixmatch})}: \mixmatchnl\ further leverages on unlabeled datasets to train the model. 
\end{itemize}
\vspace{-1mm}
Among all choices of \da\ operators, we pick and report the one with the best performance on samples
of size 1000 (since this is the labeling budget that we want to optimize under) for \da, \mixda, and \mixmatchnl.
The performance numbers reported for \bertpt\ and \bertptm\ are from the original paper \cite{posttrain}. 

\smallskip
\noindent \textbf{Roadmap}: In the remainder of this section, we first present our main result in Section~\ref{sec:exp:main} and demonstrate that our proposed solutions outperform the state-of-the-art models on all ABSA benchmark datasets; we then show a detailed comparison of the different \da\ operators, their performance,
and the improvement when we apply \mixda\ in Section~\ref{sec:exp:da}; finally, we conduct ablation analysis of the proposed \mixmatchnl\ model in Section~\ref{sec:exp:mixmatch}.

\vspace{-1mm}
\subsection{Main Results}\label{sec:exp:main}

Figure \ref{fig:expmain} shows the performance of \da, \mixda, and \mixmatchnl\ 
on the four ABSA datasets with different sizes of training data. Table \ref{tab:mainresult}
tabulates the detailed performance numbers on each dataset at size 1000 and full sizes.

\noindent \textbf{Low resource setting.} As shown in Table \ref{tab:mainresult}, 
\mixda\ and \mixmatchnl\ achieve significant
performance improvement in lower-resource settings.
In restaurant AE and ASC, \mixmatchnl\ already outperforms \bertptm, which is trained with the full data, 
using only 1,000 labeled training examples, i.e., 50\% and 28\% of training examples respectively. 
\mixda\ also achieves similar good performance with only 1000 size training set on restaurant AE and 
even outperforms \bertpt\ on full training data by 0.9\% (77.79-76.9) on the restaurant ASC task.
In laptop AE and ASC, \mixmatchnl\ achieves results
within 2.07\% and 0.04\% to \bertptm\ using
33\% or 43\% of training examples respectively. 
In general, as Figure \ref{fig:expmain} shows,
the performance gaps from the proposed methods 
(\mixda\ and \mixmatchnl) to the baseline (\bertpt) become larger 
as there are fewer labels ($<=$500).
These results indicate that the proposed methods
are able to significantly reduce the number of training labels required
for opinion mining tasks. 

\setlength{\tabcolsep}{2.7pt}
\begin{table}[!ht]
\small
\vspace{-3mm}
\caption{\small Results on 1,000 samples and full training sets.}\label{tab:mainresult}
\vspace{-3mm}
\resizebox{\linewidth}{!}{
\begin{tabular}{c|cc|cc|cc|cc}
\toprule
\multirow{2}{*}{Methods} & \multicolumn{2}{c|}{AE@1000} & \multicolumn{2}{c|}{ASC@1000} & 
\multicolumn{2}{c|}{AE@full} & \multicolumn{2}{c}{ASC@full} \\
        & rest  & laptop & rest & laptop    & rest        & laptop        & rest         & laptop        \\ \midrule
\bertpt~\cite{posttrain}     & -           & -             & -            & - & 77.97 & 84.26  & 76.90 & 75.08             \\
\bertptm\cite{posttrain}    & -           & -             & -            & - & 77.02 & 83.55  & 75.45 & 73.72        \\ \midrule
\bertfd     & 76.77 & 79.78 & 74.74 & 70.28 & 79.59 & 84.25  & 78.98 & 73.83           \\
\da         & 77.23 & 81.00 & 76.73 & 71.74 & 79.67 & \textbf{85.39}  & {79.79} & 74.02             \\
\mixda      & \textbf{77.61} & 81.19 & \textbf{77.79} & 72.72 & \textbf{79.79} & 84.07  & 79.22 & \textbf{75.34}              \\
\mixmatchnl & 77.18	& \textbf{81.48}	& 77.40	& \textbf{73.68}	& 79.65	& 85.26	& \textbf{80.45}	& 75.16 \\ \bottomrule
\end{tabular}}
\vspace*{-3mm}
\end{table}



\noindent \textbf{High resource setting.}
All three methods consistently outperform the \bertfd\ 
baseline in the high-resource setting as well
and achieve similar good performance. 
\mixmatchnl\ outperforms \bertpt\ (SOTA) in all the 4 tasks and
by up to 3.55\% (restaurant ASC).
We achieve the new SOTA results in all the 4 tasks via the combination of 
data augmentation (\da, \mixda) and SSL (\mixmatchnl).
Note that although \mixmatchnl\ does not significantly outperform
\da\ or \mixda, its models are expected to be more robust to labeling errors
because of the regularization by the unlabeled data as shown in previous SSL works \cite{DBLP:conf/iclr/MiyatoDG17,DBLP:conf/nips/CarmonRSDL19}.
This is confirmed by our error analysis where we found that most of
\mixmatchnl's mistakes are due to mislabeled test data.

We emphasize that the proposed \mixda\ and \mixmatchnl\ techniques are independent of
the underlying pre-trained LM and
we expect that our results can be further improved
by choosing a more advanced pre-trained LM or tuning the hyper-parameters more carefully.
Our first implementation of \bertfd,
which leverages a more recent BERT implementation, already
outperforms \bertpt\ and \bertptm\ but it can be further
improved.

\vspace{-1mm}
\subsection{DA operators and MixDA}\label{sec:exp:da}

We evaluate 9 \da\ operators based on the operator types introduced in Section \ref{sec:da}
combined with different pre-sampling and post-sampling strategies.
The 9 \da\ operators are listed in Table~\ref{tab:exp:da}. 
Recall that all token-level operators avoid tokens within target spans (the aspects).
When we apply an operator on a sentence $s$,
if the operator is at token-level, we apply it by $\max\{1, \lfloor |s| / 10 \rfloor\}$ times.
Span-level operators are applied one time if $s$ contains an aspect.
For ASC, we use SentiWordNet to avoid tokens conflicting with the polarity. 

\begin{table}[!ht]
\small
\vspace{-3mm}
\caption{\small Details of the 9 evaluated \da\ operators. 
For operators with TF-IDF sampling, tokens with lower TF-IDF (less important) 
are more likely to be sampled. For the \textbf{SPR} variants, all new spans are sampled from the training data. 
Similarity-based methods sample token/span with probability proportional to the similarity among the top 10 most similar tokens/spans.
BERT similarity is taken to be the cosine similarity between the [CLS] tokens' encoding.}
\vspace{-3mm}
\label{tab:exp:da}
\begin{tabular}{c c  c  c } \toprule
\textbf{Operator} & \textbf{Type} & \textbf{Pre-sampling} & \textbf{Post-sampling}\\ \midrule
\textbf{TR}        & Replace & Uniform & Word2Vec Similarity \\ 
\textbf{TR-IMP}  & Replace & TF-IDF  & Word2Vec Similarity  \\ 
\textbf{INS}       & Insert before/after & Uniform & Word2Vec Similarity \\ 
\textbf{DEL}       & Delete  & Uniform & - \\ 
\textbf{DEL-IMP} & Delete  & TF-IDF  & - \\ 
\textbf{SW}        & Swap tokens & Uniform & Uniform \\ \midrule
\textbf{SPR}       & Replace  & Uniform & Uniform \\ 
\textbf{SPR-FREQ}  & Replace  & Uniform & Frequency \\
\textbf{SPR-SIM}   & Replace  & Uniform & BERT Similarity\\
\bottomrule
\end{tabular}
\vspace*{-3mm}
\end{table}

We fine-tune the in-domain BERT model on the augmented training sets for each \da\ operator and rank these operators by their performance.
For each dataset, we rank the operators by their performance with training data of size 1000.
Table \ref{tab:darank} shows the performance of the top-5 operators and their \mixda\ version~\footnote{The \mixda\ version is generated with the \mixupnl\ hyper-parameter $\alpha$ ranging from $\{0.2, 0.5, 0.8\}$
and we report the best one.}. 

As shown in Table \ref{tab:darank}, the effectiveness of \da\ operators varies 
across different tasks. 
Span-level operators (\textbf{SPR}, \textbf{SPR-SIM}, and \textbf{SPR-FREQ}) 
are generally more effective than token-level ones in the ASC tasks. 
This matches our intuition that changing the target aspect 
(e.g., ``roast beef'' $\rightarrow$ ``vegetarian options'') is unlikely to change the sentiment on the target.
Deletion operators (\textbf{DEL} and \textbf{DEL-IMP}) perform well on the AE tasks.
One explanation is that deletion does not introduce extra information to the input sequence and
thus it is less likely to affect the target spans; but on the ASC tasks, deletion operators can remove tokens related to the sentiment on the target span.

In general, \mixda\ is more effective than \da.
Among the $36$ settings that we experimented with, we found that \mixda\ improves
the base \da\ operator's performance in $34$ ($94.4\%$) cases. 
On average, \mixda\ improves a DA operator by $1.17\%$. 
In addition, we notice that \mixda\ can have different effects on different operators
thus a sub-optimal operator can become the best choice after \mixda.
For example, in restaurant ASC, \textbf{SPR} outperforms \textbf{SPR-SIM} (the original top-1)
by 1.33\% after \mixda.

\begin{table}[!ht]
\small
\vspace*{-3mm}
\caption{\small Top-5 DA operators of each task with 1000 examples.
Recall that the baseline (\bertfd) performance is
77.26 (F1), 79.78 (F1), 74.74 (MF1), and 70.28 (MF1) on 
Restaurant-AE, Laptop-AE, Restaurant-ASC, and Laptop-ASC respectively.}\label{tab:darank}
\vspace*{-3mm}
\begin{tabular}{c|p{1.7cm}|p{1.7cm}|p{1.7cm}|p{1.7cm}} \toprule
 & \multicolumn{2}{c|}{Restaurant-AE} & \multicolumn{2}{c}{Laptop-AE} \\
Rank & \hfil Operator & \hfil \da\ / \mixda &  \hfil Operator & \hfil \da\ / \mixda \\\midrule
1    & \hfil TR  & 77.23 / 77.61 $\color{green}{\uparrow}$ & 
       \hfil DEL-IMP   & 81.00 / 81.10 $\color{green}{\uparrow}$  \\
2    & \hfil DEL  & 77.03 / 77.10 $\color{green}{\uparrow}$ & 
       \hfil SW   & 80.23 / 81.00  $\color{green}{\uparrow}$  \\
3    & \hfil SPR-SIM  & 76.88 / 76.47  $\color{red}{\downarrow}$ & 
       \hfil SPR-SIM  & 80.14 / 80.35 $\color{green}{\uparrow}$  \\
4    & \hfil TR-IMP  & 76.60 / 76.91 $\color{green}{\uparrow}$ & 
       \hfil TR-IMP   & 80.17 / 81.18 $\color{green}{\uparrow}$  \\
5    & \hfil DEL-IMP  & 76.14 / 77.09 $\color{green}{\uparrow}$ & 
       \hfil DEL   & 79.95 / 81.19 $\color{green}{\uparrow}$  \\ \bottomrule
\end{tabular}
\begin{tabular}{c|p{1.7cm}|p{1.7cm}|p{1.7cm}|p{1.7cm}} \toprule
 & \multicolumn{2}{c|}{Restaurant-ASC} & \multicolumn{2}{c}{Laptop-ASC} \\
Rank & \hfil Operator & \hfil \da\ / \mixda &  \hfil Operator & \hfil \da\ / \mixda \\\midrule
1    & \hfil SPR-SIM  & 76.73 / 76.46 $\color{red}{\downarrow}$ & 
       \hfil SPR-SIM   & 71.74 / 72.63 $\color{green}{\uparrow}$  \\
2    & \hfil SPR-FREQ  & 76.12 / 77.37 $\color{green}{\uparrow}$ & 
       \hfil SPR-FREQ   & 71.43 / 72.72 $\color{green}{\uparrow}$  \\
3    & \hfil SPR  & 75.59 / 77.79 $\color{green}{\uparrow}$ & 
       \hfil TR   & 71.01 / 71.65 $\color{green}{\uparrow}$  \\
4    & \hfil TR-IMP  & 74.42 / 74.90 $\color{green}{\uparrow}$ & 
       \hfil SPR   & 70.62 / 72.20 $\color{green}{\uparrow}$  \\
5    & \hfil INS  & 73.95 / 75.40 $\color{green}{\uparrow}$ & 
       \hfil INS   & 70.35 / 71.58 $\color{green}{\uparrow}$  \\ \bottomrule
\end{tabular}
\vspace*{-3mm}
\end{table}

\begin{figure}[!htb]
\vspace*{-2mm}
\minipage{0.23\textwidth}
  \includegraphics[width=\linewidth]{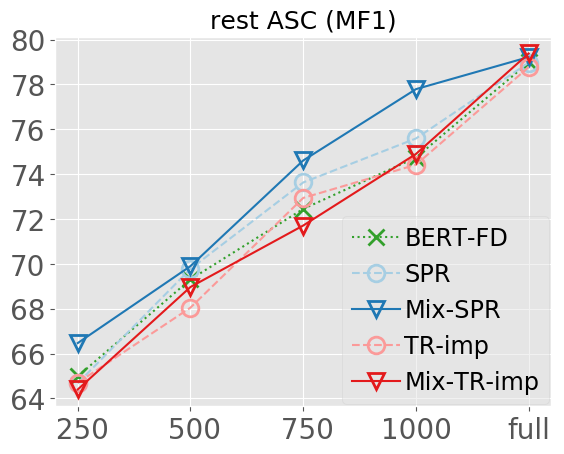}
\endminipage\hfill
\minipage{0.23\textwidth}
  \includegraphics[width=\linewidth]{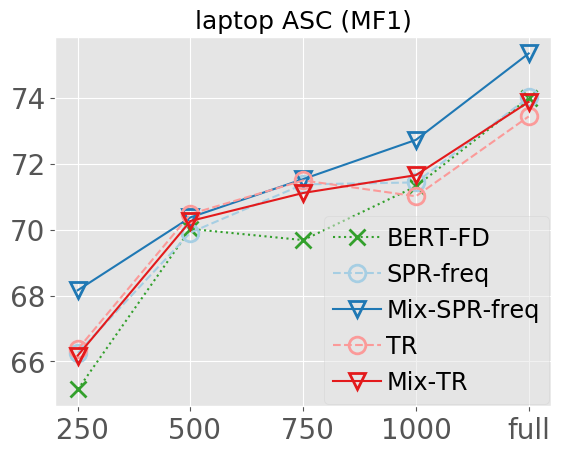}
\endminipage
\vspace*{-3mm}
\caption{\small Two representative DA operators and their MixDA versions. }\label{fig:expmixda}
\vspace*{-4mm}
\end{figure}

To verify the findings on different sizes of training data, 
we present the performance of two representative 
\da\ operators and their \mixda\ versions in the two ASC tasks at different training set sizes in Figure \ref{fig:expmixda}. The results show that
there can be a performance gap of up to 4\% among the \da\ operators and their \mixda\ versions. 
There are settings where the \da\ operator can even hurt the
performance of the fine-tuned model (restaurant@750 and laptop@1000).
In general, applying the \da\ operator with \mixda\ is beneficial. In 14/20 cases, the \mixda\ version outperforms the original operator. Note that the \mixda\ operators are optimized on datasets of size 1000, and we can achieve better results if we tune hyper-parameters of \mixda\ for each dataset size. \zhengjie{Emphasize that \mixda\ is not the optimal version} \yuliang{good.}

\vspace{-2mm}
\subsection{Ablation analysis with MixMatch$^{\text{NL}}$}\label{sec:exp:mixmatch}
We analyze the effect of each component of \mixmatchnl\ by ablation. The results are shown in Table \ref{tab:ablation}.
We consider a few variants. First, we replace the \mixda\ component with regular \da's (the ``w/o. \mixda'' row).
Second, we disable the use of unlabeled data. The resulting method is equivalent to the \bertfd\ baseline
but with \mixupnl\ as regularization (the 3rd row). 
Third, to investigate if the guessed labels by pre-mature models harm the performance,
we disable label guessing for the first 3 epochs (the 4th row).
\yuliang{added. pls check.}

\smallskip
\noindent
\textbf{Hyper-parameters. }
We tune the the hyper-parameters of \mixmatchnl\ based on our findings with \mixda.
We choose \textbf{DEL-IMP} and \textbf{SPR-FREQ} as the \da\ operators for AE and ASC respectively.
We set $(\alpha_{\text{mix}}, \alpha_{\text{aug}}, \lambda_U)$ to be $(.2, .2, .1)$ for Restaurant AE,
$(.2, .5, .1)$ for Laptop AE, and $(.8, .8, .25)$ for the two ASC datasets.
Note that training \mixmatchnl\ generally takes longer time than simple fine-tuning
thus we were not able to try all combinations exhaustively.
For \mixupnl, we pick the best result with $\alpha$ chosen from $\{0.2, 0.5, 0.8\}$.

\begin{table}[!ht]
\small
\vspace*{-3mm}
\caption{\small Ablation analysis of \mixmatchnl. 
We evaluate performance with F1 score for AE and MF1 for ASC.}\label{tab:ablation}
\vspace*{-3mm}
\begin{tabular}{c|cc|cc|cc|cc}
\toprule
\multirow{2}{*}{Methods} & 
\multicolumn{2}{c|}{AE@1000} & \multicolumn{2}{c|}{ASC@1000} & \multicolumn{2}{c|}{AE@full} & \multicolumn{2}{c}{ASC@full}  \\
        & rest  & laptop & rest & laptop    & rest        & laptop        & rest         & laptop        \\ \midrule
\mixmatchnl & 77.18	& {81.48}	& 77.40	& {73.68}	& 79.65	& 85.26	& {80.45}	& 75.16 \\
w/o. \mixda & 76.76	& 81.15	& 75.60	& 73.13	& 79.29	& 85.26	& 80.29	& 75.36  \\
\mixupnl    & 76.15 & 80.69 & 74.78 & 71.46 & 78.07 & 84.70 & 78.32 & 73.00  \\ 
w/o. pre-mature & 76.90	& 81.18	& 77.88	& 74.00	& 79.27	& 85.73	& 80.47	& 75.01 \\ \bottomrule
\end{tabular}
\vspace{-3mm}
\end{table}

\smallskip
\noindent
\textbf{Results. } Table \ref{tab:ablation} shows that both \mixda\ and unlabeled data
are important to \mixmatchnl's performance. The performance generally degrades 
as \mixda\ is removed (by up to 1.8\% in Restaurant ASC@1000)
and unlabeled data are removed (by up to 2.6\%).
The effectiveness of the two optimizations is similar among both AE and ASC tasks.
As expected, both optimizations are more effective in the settings with less data 
(a total of 9.76\% absolute improvement at size 1000 vs. 6.75\% at full size).
Finally, it is unclear whether discarding guessed labels from pre-mature models helps improve the performance
(with only $\sim$1\% difference overall).

\vspace{-1mm}
\section{Snippext in Practice} \label{sec:impact}

Next, we demonstrate \snippext's performance in practice on a real-world hotel review corpus.
This hotel review corpus consists of 842,260 reviews of 494 San Francisco hotels and is collected by an online review aggregation company whom we collaborate with. 

We apply \snippext\ to extract opinions/customer experiences from
the hotel review corpus. 
We obtain labeled training datasets from \cite{opinedb}
for tagging, pairing, and attribute classification
to train \snippext's models for the hotel domain.
In addition to their datasets, we labeled 1,500 more training examples and 
added 50,000 unlabeled sentences for semi-supervised learning.
Since the aspect sentiment data are not publicly available for the hotel corpus, 
we use the restaurant ASC dataset described in Section \ref{sec:experiment}.
A summary of the data configurations is shown in Table \ref{tab:hotel}.

We train each model as follows. 
All 4 models use the base BERT model fine-tuned on hotel reviews.
Both the tagging and pairing models are trained
using \mixmatchnl\ with the \textbf{TR-IMP} \da\ operator and 
$(\alpha_{\text{mix}}, \alpha_{\text{aug}}, \lambda_U) = (0.2, 0.8, 0.5)$.
For the attribute model, we use the baseline's fine-tuning method instead of \mixmatchnl\ 
since the task is simple and there is adequate training data available.
The sentiment model is trained with the best configuration described in the last section.

For each model, we repeat the training process 5 times and select the best performing
model in the test set for deployment. Table \ref{tab:hotel} summarizes each model's performance on various metrics.
\snippext's models consistently outperform models obtained with the baseline method in \cite{opinedb} (i.e., fine-tuned vanilla BERT) significantly. The performance improvement ranges from 
1.5\% (tagging F1) to 3.8\% (pairing accuracy) in absolute values.

\setlength{\tabcolsep}{3pt}
\begin{table}[!ht]
\small
\vspace*{-3mm}
\caption{\small Models for Hotel Extractions.}\label{tab:hotel}
\vspace*{-4mm}
\resizebox{\linewidth}{!}{
\begin{tabular}{c|cccc}\toprule
Tasks     & Train / Test / Raw & Metrics         & \snippext             & Baseline           \\ \midrule
Tagging   & 2,452  / 200  / 50,000     & P / R / F1     & \begin{tabular}{c}
71.1 / 81.0 \\
75.7
\end{tabular} & \begin{tabular}{c}
68.9 / 80.5 \\
74.2
\end{tabular} \\
Pairing   & 4,180  / 561  / 75,699     & Acc. / F1  & 84.7 / 78.3        & 80.9 / 74.5        \\
Attribute & 4,000  / 1,000 / -         & Acc. / MF1 & 88.0 / 86.9        & 86.2 / 83.3        \\
Sentiment & 3,452  / 1,120 / 35,554     & Acc. / MF1 & 87.1 / 80.7        & -       \\ \bottomrule
\end{tabular}}
\vspace*{-4mm}
\end{table}

With these 4 models deployed, \snippext\ can extract 3.49M aspect-opinion tuples 
from the review corpus, 
compared to only 3.16M tuples extracted by the baseline pipeline. 
To better understand the coverage difference,
we look into the aspect-opinion pairs extracted only by \snippext\ but \emph{not by the baseline pipeline}.
We list the most frequent ones in Table \ref{tab:hoteldiff}.
Observe that \snippext\ extracts more fine-grained opinions.
For example, ``hidden, fees'' appears 198 times in \snippext's extractions, out of 707 ``fees'' related extractions.
In contrast, there are only 124 ``fees'' related extractions with the baseline method and the most frequent ones are ``too many, fees'', which is less informative than ``hidden fees'' (and ``hidden fees'' are not extracted by the baseline method).
As another example, there are only 95 baseline extractions about the price (i.e., contains ``\$'' and a number) 
of an aspect. In comparison,  \snippext\ extracts 21,738 (228$\times$ more) tuples about the price of an aspect (e.g.,  ``\$50, parking'').

Such finer-grained opinions are useful information for various applications such as 
opinion summarization and question answering. For example, if a user asks ``Is this hotel in a good or bad location?'', then a hotel QA system can provide the general answer ``Good location'' and additionally, also provide finer-grained information to explain the answer (e.g., ``5 min away from Fisherman's Wharf'').


\begin{table}[!ht]
\small
\vspace*{-3mm}
\caption{\small{Most frequent new opinion tuples discovered by \snippext.}} \label{tab:hoteldiff}
\vspace*{-3mm}
\begin{tabular}{cc|cc} \toprule
Tuples                & Count & Tuples              & Count \\ \midrule
definitely recommend, hotel    &  1411 & own, bathroom  &  211\\ 
going on, construction  & 635 & only, valet parking  &    208\\ 
some, noise  &    532 & many good, restaurants &  199\\ 
close to, all &   449 & went off, fire alarm  &   198\\ 
great little, hotel  &    383 & hidden, fees  &   198\\ 
some, street noise    &   349 & many great, restaurants & 197\\ 
only, coffee  &   311 & excellent location, hotel   &     185\\ 
very happy with, hotel &  286 & very much enjoyed, stay & 184\\ 
\$ 50, parking &   268  & drunk, people   & 179\\ 
just off, union square &  268 & few, amenities &  171\\ 
noisy at, night & 266 & loved, staying  & 165\\ 
enjoy, city   &   245 & quiet at, night & 163\\ 
hidden, review &  227 & some, construction   &    161\\ 
definitely recommend, place   &   217 & some, homeless people  &  151\\ 
too much trouble, nothing    &    212 & truly enjoyed, stay  &  145 \\ \bottomrule
\end{tabular}
\vspace*{-1mm}
\end{table}

\vspace{-1mm}
\section{Related Work} \label{sec:related}



Structured information, such as aspects, opinions, and sentiments,
which are extracted from reviews are used to support 
a variety of real-world applications~\cite{archak2007show, kim2011comprehensive, marrese2014novel, opinedb, voyageur}.
Mining such information is challenging and 
there has been extensive research on these topics~\cite{kim2011comprehensive, liu2012sentiment}, 
from document-level sentiment classification 
\cite{zhang2015character,maas2011learning} to
the more informative Aspect-Based Sentiment Analysis (ABSA) \cite{pontiki2014semeval, pontiki2015semeval} or 
Targeted ABSA \cite{sentihood}. 
Many techniques have been proposed for review mining, from
lexicon-based and rule-based approaches~\cite{hu2004mining, ku2006opinion, poria2014rule} to
supervised learning-based approaches~\cite{kim2011comprehensive}.
Traditionally, supervised learning-based approaches~\cite{jakob2010extracting, zhang2011extracting, mitchell2013open} 
mainly rely on Conditional Random Fields (CRF) and require heavy feature engineering.
More recently, deep learning models~\cite{tang2015document, poria2016aspect, wang2017coupled, xu2018double} 
and word embedding techniques
have also been shown to be very effective in ABSA tasks even with little or no feature engineering. 
Furthermore,
the performance of deep learning approaches~\cite{posttrain, auxiliarysentence} 
can be further boosted by pre-trained language models, 
such as BERT \cite{bert} and XLNet \cite{xlnet}.

\snippext\ also leverages deep learning and pre-trained LMs to perform the review-mining-related tasks and focuses on the problem of reducing the amount of training data required. One of its strategies is to augment the available training data through data augmentation.
The most popular \da\ operator in NLP is by replacing words with other words selected by random sampling~\cite{eda}, 
synonym dictionary~\cite{zhang2015character}, semantic similarity~\cite{wang2015s}, 
contextual information~\cite{kobayashi2018contextual}, and 
frequency~\cite{fadaee2017data, xie2019unsupervised}. 
Other operators, such as random insert/delete/swap words~\cite{eda} and back translation~\cite{yu2018qanet}, are also proved to be effective in text classification tasks. 
As naively applying these operators may produce significant distortion to the labeled data, \snippext\ proposes a set of $\da$ operators suitable for opinion mining and only ``partially'' augments the data through \mixda.



A common strategy in Semi-Supervised Learning (SSL) is \\
Expectation-Maximization (EM)~\cite{nigam2000text}, 
which uses both labeled and unlabeled data 
to estimate parameters in a generative classifier, such as naive Bayes.
Other strategies include self-training~\cite{rosenberg2005semi, wang2008semi, sajjadi2016regularization},
which first learns an initial model from the labeled data then uses unlabeled data to further teach and learn from itself, 
and multi-view training~\cite{blum1998combining, zhou2005tri, xu2013survey, clark2018semi}, 
which extends self-training to multiple classifiers that teach and learn from each other
while learning from different slices of the unlabeled data.
MixMatch \cite{mixmatch,remixmatch,mma} is a recently proposed SSL paradigm that
extends previous self-training methods by interpolating
labeled and unlabeled data. MixMatch outperformed previous SSL algorithms and 
achieved promising results in multiple image classification tasks with only few hundreds of labels. 
\snippext\ uses \mixmatchnl, an adaptation of MixMatch to the text setting. 
\mixmatchnl\ demonstrated SOTA results in many cases and this
opens up new opportunities for 
leveraging the abundance of unlabeled reviews that are available on the Web.
In addition to pre-training word embeddings or LMs, 
the unlabeled reviews can also benefit fine-tuning of LMs in obtaining
more robust and generalized models \cite{DBLP:conf/nips/CarmonRSDL19, DBLP:conf/iclr/MiyatoDG17,hendrycks2019augmix}.
\zhengjie{Citation added}

\vspace{-1mm}
\section{Conclusion} \label{sec:conclusion}

We proposed \snippext, a semi-supervised opinion mining system that extracts aspects, opinions, and sentiments from text.
Driven by the novel data augmentation technique \mixda\ and semi-supervised learning algorithm \mixmatchnl,
\snippext\ achieves SOTA results in multiple opinion mining tasks with only half the amount of training data used by SOTA techniques.
\snippext\ is already making practical impacts on our ongoing collaboration with a hotel review aggregation platform
and a job-seeking company. In the future, we will explore optimization opportunities
such as multitask learning and active learning to further reduce the labeled data requirements in \snippext.

\bibliographystyle{ACM-Reference-Format}
\bibliography{main}

\appendix

\end{document}